\documentclass[12pt]{article}
\usepackage{cite}
\usepackage{amsmath,amssymb,amsfonts}
\usepackage{algorithmic}
\usepackage{graphicx}
\usepackage{textcomp}
\usepackage{xcolor}
\usepackage{geometry}
\usepackage{setspace}
\usepackage{authblk}

\def\BibTeX{{\rm B\kern-.05em{\sc i\kern-.025em b}\kern-.08em
    T\kern-.1667em\lower.7ex\hbox{E}\kern-.125emX}}

\begin{document}

\title{Report Title\\
}

\title{Literature Review}

\author[1]{Jiaxin Xu\thanks{Jiaixn Xu: jiaxin7@ualberta.ca}}
\author[1]{Rui Wang\thanks{Rui Wang: rw2@ualberta.ca}}
\author[1]{Vaibhav Rakheja\thanks{Vaibhav Rakheja: rakheja@ualberta.ca}}
\affil{department of Computing Science, Multimedia,University of Alberta, Edmonton, Canada}

\date{}

\maketitle
\begin{abstract}
Our research topic is Human segmentation with static camera. This topic can be divided into three sub-tasks, which are object detection, instance identification and segmentation. These sub-tasks are three closely related subjects. The development of each subject has great impact on the other two fields. In this literature review, we will first introduce the background of human segmentation and then talk about issues related to the above three fields as well as how they interact with each other.  
\end{abstract}

\section{Background}
\label{background}
Topics related to human segmentation have been increasingly active, due to the high demand of real-life applications, such as video surveillance, virtual-reality simulation, action localization, and 3D human modeling \cite{berretti2018representation}. Thanks to the advancement in object detection, instance recognition and semantic segmentation, human segmentation has become easier and more feasible.

The goal of human segmentation is to identify a human in an image or video and separate it from the background. In terms of image, background and foreground are static while for static camera, the foreground and background changes over time. In terms of moving camera, it is even harder to identify foreground and background due to the dynamic change of each pixel. So methods deal with these three situations differ a lot, for image we can apply background subtraction techniques and for videos we have to measure the change of each pixel. For example, Elnagar et al. proposed a background constraint motion detection method which maps pixels in successive images to deal with moving camera segmentation\cite{elnagar1995motion} and Yin et al. presented a frame differences fusion method for face tracking with moving camera\cite{yin1999integrating}. In this literature review we will mainly focus on human segmentation with static camera.

Traditional segmentation methods can be roughly divided into two categories, top-down methods, like \cite{yuille1992feature}\cite{borenstein2002class}\cite{yin2001nose} and bottom-up methods, like\cite{carreira2012object}\cite{arbelaez2012semantic}. Top-down methods will first try to extract representation of an object, such shapes, appearance characteristic and texture and then use these prior knowledge to guide segmentation process. The drawback is there are a variety of shapes and appearances for objects, even though they belong to the same category. Bottom-up methods will first generate candidate regions which may contain an object and then identify these regions with continuity of gray-level, texture and bounding contours. Therefore, the performance of bottom-up methods is highly depended on the accuracy of candidate regions. There are also methods try to unify top-down and bottom-up methods aim to attain mutual complementary.

\section{Terminology}
\label{terminology}
\subsection{Object Detection}
Object detection deals with instance detection of a specific class in digital images or videos, such as human, face and animals. Previous methods like \cite{xu2012edge} and \cite{yin2004scalable} use edge detection skills to assist object detection process. Moreover, Singh et al. introduced new weighting functions and integrated weighting parameters into edge characteristic of an image\cite{singh2005gaussian}. In \cite{patel2018erel}, Extremal Regions of Extremum Levels (EREL) was combined with convolutional neural network to detect defect in bottle manufacturing process. EREL was also applied to extract luminal area of human
coronary for segmentation of arterial wall boundaries from Intravascular Ultrasound (IVUS) images\cite{li2018erel}. Object Detection not only has medical applications \cite{mukherjee2018atlas}\cite{soltaninejad2018towards}, but is also developed for airport ground traffic control\cite{liu2018synthetic}. 

In recent years, object detection has been dominated by the deep convolutional neural network, for example, Fast R-CNN\cite{girshick2015fast}, Faster R-CNN\cite{ren2015faster} and YOLO\cite{redmon2016you}. 

Fast R-CNN is an object detection method with a backbone of deep VGG16 network\cite{simonyan2014very}. Different from R-CNN which uses selective search to implement region proposal, Fast R-CNN speeds up the whole framework by applying sharing computation with Spatial pyramid pooling networks(SPPnets)\cite{he2015spatial}. The speed of R-CNN, which runs at about 47 seconds per image has been improved by 10x at the testing phase. The accuracy also increased from 53.7\% to 66\% mAP on PASCAL VOC. Faster R-CNN further refined the framework with Region Proposal Network (RPN) instead of a selective search to generate region proposals and performing at a rate of 5 fps on a GPU. Later on, the appearance of YOLO has broken the record with an achievement of 45 fps and double mAP results comparing with other detection methods. YOLO frame detection as a regression problem and will only look at an image once to process recognition. They take advantage of 24 convolutional network layers to extract features of an image while the following two fully connected layers are used to output probabilities and coordinates.

Object detection is closely related to instance segmentation. During the instance segmentation process, the method should be able to detect an object and tell which class it belongs to and finally separate it from the image background. Sometimes, segmentation information can also, in turn, facilitate the result of object detection.

\subsection{Semantic Segmentation}
Semantic segmentation refers to the process of classifying each pixel with a class label. It can be seen as a pixel level image classification. Semantic segmentation plays an important role in human-computer interaction, action localization and robotic field.

In 2016, Long et al. proposed Fully Convolutional Network(FCN) for Semantic Segmentation\cite{long2015fully}. This network can efficiently produce dense predictions for each pixel from arbitrary-sized input and draw the connection to prior models, such as AlexNet, VGGNet and so on. By framing the above models into fully convolutional networks and leveraging their pre-trained representation result with fine-tuning, FCN achieves great segmentation performance. In 2017, a semantic segmentation method based on Densely Connected Convolutional Networks(DenseNets)\cite{huang2017densely} is presented by Jégou et al\cite{jegou2017one}. They combined FCN and DenseNet to deal with semantic segmentation and achieved a good result on urban scene benchmark datasets. Furthermore, FCN was also applied to ultrasound image segmentation\cite{yang2018ivus}. Multi-Scale Context Aggregation by Dilated Convolutions is a semantic segmentation module used to systematically aggregate multi-scale contextual information\cite{yu2015multi}. Their module makes use of Dilated Residual Networks\cite{yu2017dilated} can enhance segmentation accuracy without losing resolution.

Semantic segmentation can be used to realize instance segmentation and there are a variety of modules to attain comparable semantic segmentation results. The difference between semantic segmentation and instance segmentation is that the former is used to implement group instance segmentation while the latter can achieve separate instance segmentation. That means, if there are a group of people in an input image, instance segmentation can tell how many people are there while semantic segmentation cannot.

\subsection{Instance Segmentation}
There are three orders of execution for instance segmentation, which are segmentation-first, instance-first or implement the two processes simultaneously. 

In 2014, Hariharan and his groups presented a segmentation-first instance segmentation methods, called Simultaneous Detection and Segmentation\cite{hariharan2014simultaneous}. They first implement proposal generation with Multi-scale Combinatorial Grouping(MCG)\cite{arbelaez2014multiscale} to produce 2000 region candidates each image. Then, extract features using a convolutional network (ConvNet) and classification by Support Vector Machine (SVM). Kirillov et al. proposed a new modelling paradigm for instance-aware semantic segmentation, named InstanceCut\cite{kirillov2017instancecut}. They implement an instance-agnostic semantic segmentation with standard ConvNet and extract instance-boundaries with a new instance-aware edge detection model. There is an instance-first method, presented by Dai et al.\cite{dai2016instance}. Their method consists of instance identification, mask estimation and classification formed a cascade structure. Fully convolutional instance-aware semantic segmentation\cite{li2017fully} is a special approach very close to simultaneously segmentation, which performs instance masks prediction and classification jointly.

The blossom of instance segmentation enables human segmentation, animal segmentation and many other challenging tasks. In terms of human segmentation, we will introduce several outstanding methods in the following section.

\section{Human Segmentation}
\label{RW}
One of the popular topics in instance segmentation is human segmentation. Detecting and segmenting human images are challenging due to the variety of human shapes and appearances. There are many approaches to human segmentation. Some earlier research papers focus on extracting distinctive features from an image and the extracted features are used for object classification and segmentation, such as Histograms of Oriented Gradient (HOG) descriptors\cite{dalal2005histograms} and Scale Invariant Feature Transform (SIFT)\cite{lowe2004distinctive}. Later, the pose-based human segmentation approach like \cite{singh2008human}, \cite{singh2005pose} and \cite{zhang2019pose2seg} are proposed.  In \cite{zhang2019pose2seg} Zhang et al. detects human body key parts, and then builds up a segmentation mask on top of the key parts. Up to now, the most accurate human segmentation approach first identifies the region or bounding box around the human image then perform segmentation, one of the examples is Mask R-CNN\cite{he2017mask}. 

In the early works, hand-engineered features were computed over an entire image for human detection and human pose estimation. In 2005,  
the Radon transform based action recognition method was developed to assist ground air traffic control\cite{singh2005visual}. Radon transform was used to combat disadvantage in Hough transform and was used first time for human gesture recognition. The information generated using Radon transformation is used to classify the gestures in air traffic control. Many hand gesture recognition methods extract distinctive features then classify gestures\cite{wang2018survey}.  

\begin{figure}[htbp]
\centerline{\includegraphics[width=6in]{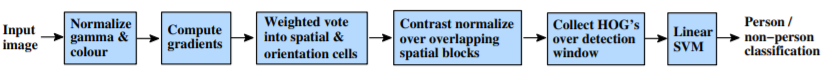}}
\caption{An overview of feature extraction and object classification using Histogram of Gradients descriptors. Reprinted from Navneet Dalal and Bill Triggs, “Histograms of oriented gradients for human detection,”2005}
\label{fig1}
\end{figure}

The other two feature extraction methods are Histograms of Oriented Gradient (HOG) descriptors and Scale Invariant Feature Transform (SIFT). In 2005, Navneet Dalal and Bill Triggs\cite{dalal2005histograms} proposed HOG descriptors for human detection. The overview of HOG feature extraction and object detection chain is illustrated in Figure~\ref{fig1}. HOG is a global feature descriptor that looks at the image as a whole for human detection. It characterizes object shape and appearance by the distribution of local intensity gradients or edge directions. HOG first computes centred horizontal and vertical gradients, then calculates gradients orientation and magnitude. For the input video frames, HOG divides an image into blocks with a 50 percent overlap. Later, HOG interpolates gradient orientation in each block into 9 bins. The HOG achieves very good results for human detection. In this paper, Navneet Dalal and Bill Triggs\cite{dalal2005histograms} combined HOG with SVM classifier to output person/non-person classification.

\begin{figure}[htbp]
\centerline{\includegraphics[width=5in]{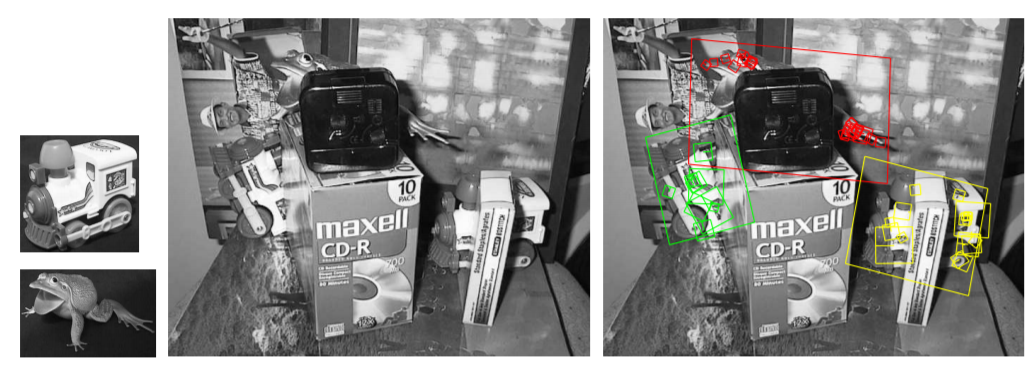}}
\caption{An example of object detection in a cluttered and occluded image using SIFT. Reprinted from David G Lowe, “Distinctive image features from scale-invariant keypoints,”International journal of computer vision, vol. 60, no. 2, pp. 91–110, 2004.}
\label{fig2}
\end{figure}

The other approach, SIFT, computes a set of unique image features that are invariant to scaling, rotation, change in illumination, and is not disrupted by occlusion\cite{lowe2004distinctive}. To obtain the set of image features, SIFT first looks over the entire image using a difference-of-Gaussian function to identify points in the image that are not changing according to image scale and orientation. A detailed model is applied to find the location and scale of identified keypoints with high stability. Each keypoint is assigned with one or more orientations based on local image gradient directions. All future operations on images are applied relative to assigned orientation, scale and location, making local features invariant to transformation. Lastly, local image gradients around keypoints are transformed into representations that allow distortion and change in illumination. Figure ~\ref{fig2} demonstrates SIFT's performance on recognizing objects in a cluttered and occluded image.   

There are drawbacks to the HOG and SIFT, such as missing objection and misidentifying objects. Due to these drawbacks, more approaches based on deep learning networks are developed to increase accuracy in detection results. 

The research on pedestrian detection achieves very good results and has many real-life applications. In \cite{mao2017can}, the paper integrates semantic channel features and heatmap channels are integrated into CNN-based pedestrian detectors. Two feature channels improve detection and localization accuracy. The paper also proposed a framework called HyperLearner that learns channel features and pedestrian detection at the same time.

In pose-based human segmentation, multi-person pose estimation has been a very popular topic and there is a lot of progress. For example, Chen et al. \cite{chen2018cascaded} proposed a Cascade Pyramid Network. The framework has two stages: GlobalNet and RefineNet. GlobalNet locates keypoints based on Feature Pyramid Network to identify keypoints that have distinctive different features. For keypoints that require more context information to localize, RefineNet is applied. RefineNet integrates resulted features from feature pyramid network and online hard keypoint mining to locate keypoints that GlobalNet is unable to identify. The Feature Pyramid Network (FPN) in GlobalNet was originated from\cite{lin2017feature}. The FPN includes a bottom-up pathway, a top-down pathway, and lateral connections. It introduces a new type of feature pyramids in convolutional neural networks that outperforms the state-of-art feature extraction framework in speed and accuracy.

\begin{figure}[htbp]
\centerline{\includegraphics[width=5in]{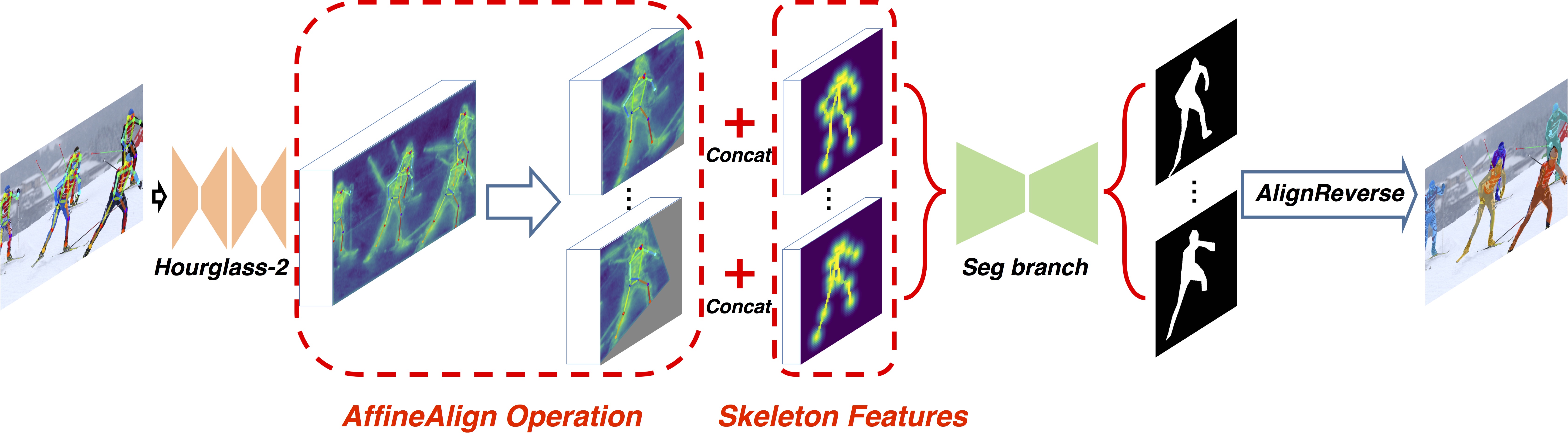}}
\caption{An overview structure of Pose2Seg proposed by Song-Hai Zhang, Ruilong Li, Xin Dong, Paul Rosin, Zixi Cai, Xi Han, Dingcheng Yang, Haozhi Huang, and Shi-Min Hu, “Pose2seg: Detection free human instance segmentation,” in Proceedings of the IEEE Conference on Computer Vision and Pattern Recognition, 2019, pp. 889–898}
\label{fig3}
\end{figure}

The other pose-based human segmentation method was proposed by Zhang et al.\cite{zhang2019pose2seg}. Figure ~\ref{fig3} illustrated the overall structure of the pose-based segmentation. The framework first applies a base layer to perform feature extraction, then uses an align model called Affine-Align to align the feature map with human poses. Additionally, skeleton features are linked with Affine-Align Operation, which is comprised of part affinity fields that represent the skeleton structure of human pose and part confidence map that highlights the region around boy keypoints. Lastly, the segmentation model named SegModule is applied. SegModule is composed of convolutional layers, residual units, upsampling layers, stride layers. The SegModule creates final masks of human segmentation. The part affinity fields in pose-based human segmentation were first introduced in \cite{cao2017realtime}. Part affinity fields are a set of 2-D vectors that contain location and orientation information of limbs within the image. The paper also developed a 2-D framework that detects the confidence map and predicts affinity fields at the same time. One of the feature set that can be used in pose estimation is OpenPose\cite{heath2018detecting}.

Papandreou et al. addressed the multi-person pose estimation framework\cite{papandreou2017towards}. In the paper, Papandreou et al. developed a two-stage model: in the first stage, Faster R-CNN is employed to detect regions that possibly contain human images; in the second stage, the framework use ResNet to predict locations of keypoints in each region that is likely to contain a person. This work resolve person detection in clustered images. 

Apart from the pose-based human segmentation method, there is another approach that combines human detection and segmentation. In 2015, Pinheiro et al. proposed a model based on the discriminative convolutional network\cite{pinheiro2015learning}. The model outputs a class-agnostic segmentation mask while predicts the probability of the mask contains an object. The two tasks, mask computation and probability prediction, share most layers of convolutional neural networks, and only the last layers for two task are different. Sharing layers of the network reduces model capacity and computational complexity. 

\begin{figure}[htbp]
\centerline{\includegraphics[width=4in]{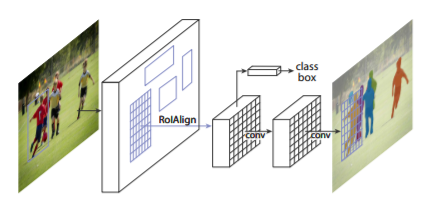}}
\caption{An overview of Mask R-CNN proposed by Kaiming He, Georgia Gkioxari, Piotr Dollar, and  Ross  Girshick,   “Mask  r-cnn,” in Proceedings of the IEEE international conference on computer vision, 2017, pp. 2961–2969}
\label{fig4}
\end{figure}

In contrast to object segmentation methods developed by Pinheiro et al. \cite{pinheiro2015learning} where classification is dependent on mask segmentation, the most state-of-art techniques, Mask R-CNN\cite{he2017mask} simultaneously detect and segment mask at high accuracy. The Mask R-CNN pipeline is included in ~\ref{fig4} For object detection, the Mask R-CNN uses ResNet 101 to extract features from the images. The feature maps are input to Regional Proposal Networks which predicts if a human is present in
the region and outputs the regions that contain humans. To get the universal size of proposal regions, a pooling layer converts all regions to the same shape. Later, these regions are passed
through a fully connected network and class labels and bounding boxes are produced. The object segmentation part of Mask R-CNN further segments the region of interest. It calculates Intersection over Union (IoU) with ground truth box. Once the region of interest is generated, Mask R-CNN uses fully convolutional networks (FCN) to create a mask on the intersection between the region of interest and ground truth image. The FCN avoids vector transformation that loses spatial dimensions. It operates on the input image and creates corresponding output with spatial information. To predict pixel-accurate masks, ROIAlign is applied to align feature maps of the region of interest. The disadvantage of Mask R-CNN is that the segmentation results are heavily dependent on detection performance. For example, the poor performance of detection on overlapped human images will negatively impact segmentation outputs. To find the optimal trade-off between accuracy and speed in the object detection phase, Huang et al. compared state-of-art CNN-based object detectors\cite{huang2017speed} in 2017. As a result, R-FCN and SSD models are faster while Faster R-CNN is slightly slower but achieves higher accuracy. The computational speed of Faster R-CNN can be improved with a reduced number of regional proposals from 300 to 50.

\section{Conclusion}
\label{conclusion}

Human segmentation has many real-life applications and is very important in video surveillance such as action recognition, suspicious activity detection, and activity tracking. Furthermore, it is a fundamental step in 3-D model reconstruction. 

This report introduces the background of human segmentation and extensively reviews previous researches on three sub-task in human segmentation: object detection, instance identification and segmentation.

For future improvement, our team also review the Generative adversarial networks (GANs) which could introduce noisy images/videos into our project to improve our framework accuracy\cite{singh2018step} \cite{eusebio2018semi}. Due to the time constraints, we will include GANs as the future direction of our project.


\bibliographystyle{IEEEbib}
\bibliography{egbib}

\end{document}